\pdfoutput=1
\documentclass[11pt]{article}
\usepackage[final]{acl}
\usepackage{listings}
\usepackage{xcolor}
\usepackage{times}
\usepackage{latexsym}
\usepackage{amsmath}
\usepackage{amssymb}
\usepackage{subfigure}
\usepackage{enumitem}
\usepackage{pifont}
\usepackage{amsthm}

\usepackage[T1]{fontenc}
\usepackage[utf8]{inputenc}
\usepackage{microtype}
\usepackage{inconsolata}
\usepackage{graphicx}
\usepackage[breaklinks]{hyperref}
\usepackage{breakurl}
\usepackage{multirow}
\usepackage{graphicx}
\usepackage{wrapfig}
\usepackage{booktabs}
\usepackage{changepage}
\usepackage{makecell}
\usepackage{setspace}
\title{AgentRM: Enhancing Agent Generalization with Reward Modeling}
\author{
 \textbf{Yu Xia\textsuperscript{1}},
 \textbf{Jingru Fan\textsuperscript{1}},
 \textbf{Weize Chen\textsuperscript{1}},
 \textbf{Siyu Yan\textsuperscript{1}},
 \textbf{Xin Cong\textsuperscript{1}},
 \textbf{Zhong Zhang\textsuperscript{1}},
 \textbf{Yaxi Lu\textsuperscript{1}},
 \\
 \textbf{Yankai Lin\textsuperscript{2}\thanks{Corresponding authors.}},
 \textbf{Zhiyuan Liu\textsuperscript{1}\footnotemark[1]},
 \textbf{Maosong Sun\textsuperscript{1}}
 \\
 \textsuperscript{1}Department of Computer Science and Technology, Tsinghua University \\
 \textsuperscript{2}Gaoling School of Artificial Intelligence, Renmin University of China\\
\texttt{xiayu23@mails.tsinghua.edu.cn, liuzy@tsinghua.edu.cn}
}

\begin{document}
\maketitle
\begin{abstract}
Existing LLM-based agents have achieved strong performance on held-in tasks, but their generalizability to unseen tasks remains poor. Hence, some recent work focus on fine-tuning the policy model with more diverse tasks to improve the generalizability. In this work, 
we find that finetuning a reward model to guide the policy model is more robust than directly finetuning the policy model.
Based on this finding, we propose AgentRM, a 8B generalizable reward model, to guide the policy model for effective test-time search.
We comprehensively investigate three approaches to construct the reward model, including explicit reward modeling, implicit reward modeling and LLM-as-a-judge.
We then use AgentRM to guide the answer generation with Best-of-N sampling and beam search.
We show that AgentRM is robust to paraphrasings of task instructions and can generalize to unseen tasks that require novel optimal behavior.
Through extensive evaluation across nine tasks spanning four categories, AgentRM enhances the non-finetuned 8B policy model by $8.8$ points on average, surpassing the top general agent by $4.0$.
Moreover, it demonstrates weak-to-strong generalization, yielding greater improvement on more powerful policy models.
As for the specializability, AgentRM can also boost a finetuned policy model and outperform the top specialized agent by $11.4$ on three held-in tasks.
Further analysis verifies its effectiveness in test-time scaling.
We release the code and data at \href{https://github.com/thunlp/AgentRM}{https://github.com/thunlp/AgentRM}.

\end{abstract}

\section{Introduction}
\label{sec:intro}
Large language model (LLM)-based agents~\cite{mialon2023augmented, sumers2023cognitive} have become a promising solution to complex interactive tasks~\cite{xi2024agentgym} in recent years.
While specialized agents~\cite{wang2024learning, qin2023toolllm} achieve strong performance on held-in tasks, their generalizability to unseen tasks is poor.
To address this challenge, existing works focus on integrating more diverse agent tasks including human-crafted~\cite{zeng2023agenttuning, chen2024agent, xi2024agentgym, zhang2024agentohana, acikgoz2025singlemodelmastermultiturn, zhuang2025hephaestus} and LLM synthesized~\cite{hu2024agentgen,fu2025agentrefineenhancingagentgeneralization}, to perform multi-task fine-tuning on the base LLM.
\begin{figure}[t]
      \includegraphics[width=\columnwidth]{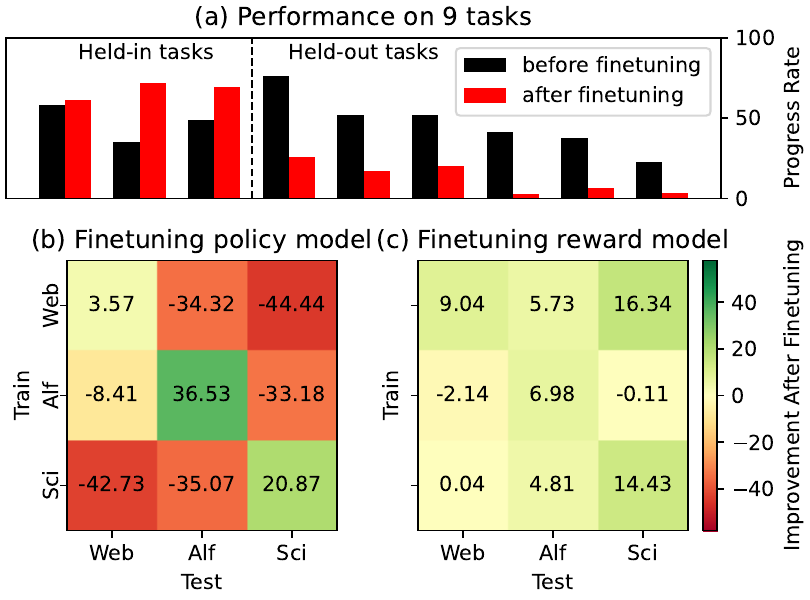}
      \caption{Finetuning the reward model is more robust than finetuning the policy model for agent tasks. (a) Finetuning the policy model leads to severe degradation on held-out tasks. (b)(c) show the performance of Best-of-5 with a reward model. Finetuning the policy model on one task degrades on others while finetuning the reward model mostly generalized to others.}
      \label{fig:motivation}
\end{figure}

However, we find simply finetuning the base LLM improves held-in task performance but degrades held-out task performance (Figure~\ref{fig:motivation}(a)). This suggests that finetuning—which directly adapts the policy model responsible for token-by-token action generation—may overfit to seen action sequences, while suppressing unseen but potentially valid action sequences. To mitigate this distribution shift and improve robustness for unseen tasks, we hypothesize that finetuning a separate reward model to guide the policy model is more effective. This approach leverages the inherent advantage of reward modeling: its regression-based training objective focuses on learning task-level value functions, making it less sensitive to shifts in the specific distribution of generated action tokens compared to direct policy optimization. To test this hypothesis in our preliminary experiment, we perform Best-of-5, i.e. generating 5 candidate trajectories with the policy model and selecting one using the reward model.
Figure~\ref{fig:motivation}(b)/(c) shows the improvement after fine-tuning the policy/reward model respectively on individual tasks.
In Figure~\ref{fig:motivation}(b), only the diagonal values, i.e. performance of the held-in task which is seen during training, are positive.
Contrastly, Figure~\ref{fig:motivation}(c) reveals predominantly positive values, indicating that finetuning the reward model on a single task can enhance the performance on unseen tasks.

Inspired by this, we introduce AgentRM, a generalizable reward model, to guide the policy model for effective test-time search.
Since the effective construction of the reward model for agent tasks remains an open question due to environment dynamics and long-horizon decision-making challenges,
we investigate three representative reward modeling approaches including (1) explicit reward modeling \cite{zhang2024rest} which learns the step-level rewards annotated by tree search, (2) implicit reward modeling \cite{yuan2024freeprocessrewardsprocess} which derives the inherent step-level rewards by training on outcome rewards, and (3) LLM-as-a-judge \cite{zheng2023judging} which directly prompts an LLM to assess the agent trajectory.
These methods represent a progressive reduction in workload for both reward model training data construction and model training.
We then use AgentRM to guide the answer generation in the Best-of-N sampling and step-level beam search. 

Extensive experiments demonstrate that explicit modeling consistently achieves the most significant improvements across nine agent tasks, including web navigation, embodied planning, text games, and tool usage.
Concretely, it enhances the base policy model by 8.8 points, surpassing the top general agent by 4 points.
Moreover, our reward model trained on states sampled by LLaMA-3-8B can be directly applied to other policy models in a plug-and-play manner, yielding greater improvement of 12.6 points on LLaMA-3-70B.
As for the specializability, it can also boost a finetuned policy model, surpassing the top task-specific agent by 11.4 points.
Further analysis including the scaling trend of training data, ablation on state representation and test-time scaling highlights the scalability of AgentRM.
\begin{figure*}[t]
    \centering
    \includegraphics[width=1\textwidth]{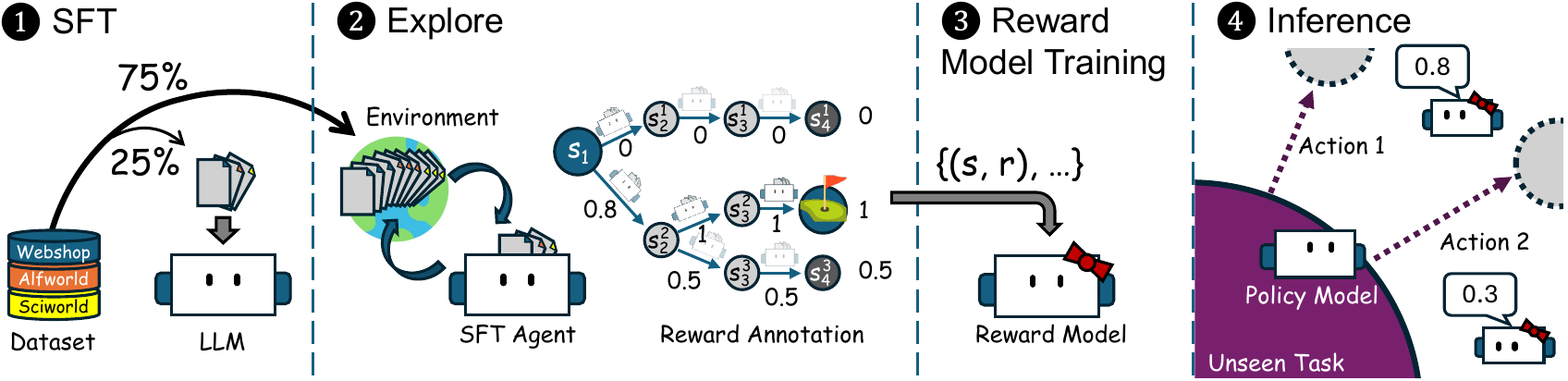}
    \caption{Overview. \ding{182} Deriving a supervised fine-tuned (SFT) agent on expert trajectories. \ding{183} Constructing search trees by exploring the environment using the SFT agent. \ding{184} Training a generalizable reward model, on state-reward pairs extracted from search trees. \ding{185} Enhancing the policy model, regardless of its initial strength, through test-time search guided by our reward model for unseen tasks such as embodied planning, text game, tool using etc.}
  \label{fig:overview}
\end{figure*}
\section{Task Formulation}
The agent task with environment feedback can be formalized as a partially observable Markov decision process 
$(\mathcal{U},\mathcal{S},\mathcal{A},\mathcal{O},\mathcal{T},\mathcal{R})$ with instruction space $\mathcal{U}$, state space $\mathcal{S}$, 
action space $\mathcal{A}$, observation space $\mathcal{O}$, state transition function
$\mathcal{T}:\mathcal{S}\times\mathcal{A}\to\mathcal{S}$, and reward function $\mathcal{R}:\mathcal{S}\times\mathcal{A}\to[0, 1]$. 
The initial state $s_1=(u, o_0)\in\mathcal{S}$ consists of task instruction $u$ and the initial observation $o_0$.
At step $t$, conditioned on the current state $s_t$, the agent generates the next action $a_{t}\sim\pi(\cdot|s_t)$ based on its policy $\pi$.
Then, the agent receives the environment observation $o_t\in\mathcal{O}$ and the state transforms to $s_{t+1}=(s_{t},a_{t},o_{t})=(u,o_0,a_{<t+1},o_{<t+1})$ according to transition function $\mathcal{T}$.
The agent continues to interact with the environment until the task is finished or the maximum step is reached.
The environment only provides the outcome reward at the final step $r_T(s_T,a_T)\in\mathcal{R}$, where $T$ denotes the total step number.
As illustrated in Section~\ref{sec:rm}, we train a process reward model that produces rewards for intermediate steps $r_t(s_t,a_t), t<T$.
We discuss the training details in Section~\ref{sec:rm}.
\section{Methodology}
The overview is depicted in Figure~\ref{fig:overview}.
Section~\ref{sec:bc} describes the behavior cloning through which we derive a policy model with basic task ability on held-in tasks.
Section~\ref{sec:rm} elaborates on how we use the derived policy model to build our generalizable reward model.
Section~\ref{sec:guide} explains how we use our reward model to enhance the policy model's decision-making ability through test-time search.

\subsection{Behavior Cloning}
\label{sec:bc}
Behavior cloning serves as an optional preliminary step in our framework. While it is permissible to employ a non-supervised fine-tuned (non-SFT) agent for trajectory sampling in the subsequent stage, this approach requires the agent to possess sufficient performance capabilities to ensure effective exploration. In this work, we use a SFT 8B alternative to a more costly and performant 70B model \footnote{The SFT LLaMA-3-8B achieves 61.4, 71.6, 66.6 on Webshop, Alfworld, Sciworld respectively. The LLaMA-3-70B achieves slightly higher performance of 63.4, 73.3, 77.1.}.
To obtain an initial policy $\pi_{init}$ with basic task ability, crucial for collecting high-quality states, we split a portion of task instructions from the training set, annotate them by an expert agent and conduct supervised fine-tuning (SFT) on the expert trajectories $D_{expert}=\{(u^i,o_0^i,a_t^{i},o_t^i)_{t=1}^{T_i}\}_{i=1}^{N}$ as follows:
\begin{equation}
    \mathcal{L}(\theta)=-\sum_{i=1}^N\sum_{t=1}^{T_i}\log\pi_\theta(a_t^i\mid u^i, o_0^i, a_{<t}^i,o_{<t}^i)
\end{equation}
where $\theta$ denotes the parameters of the policy model, $N$ denotes the number of trajectories in $D_{expert}$, $T_i$ denotes the total step of the $i$-th trajectory.

\subsection{Reward Modeling}
\label{sec:rm}
Since the effective construction of reward models in agent tasks remains underexplored, we investigate three methods with different emphases.
Explicit reward modeling (Section~\ref{sec:explicit}) employs tree search for automatic process reward annotation, distributing the sparse outcome rewards to each step in an interpretable way.
Implicit reward modeling (Section~\ref{sec:implicit}) eliminates the annotation of step-level reward and learns it implicitly.
LLM-as-a-judge (Section~\ref{sec:llmasajudge}) is a training-free method relying on the general judging ability of the LLM.

\subsubsection{Explicit Reward Modeling}
\label{sec:explicit}
Explicit reward modeling typically defines process reward as Q-value \cite{watkins1992q}, i.e. expected accumulated rewards starting from a state, and calculates it by Monte Carlo estimation on random rollouts.
Given that agent tasks typically involve long-chain reasoning and vast search space, we organize the agent's search trajectories into tree structures and employ a Monte Carlo Tree Search (MCTS)-inspired algorithm to avoid redundant exploration while encouraging sampling diversity.

The search tree consists of nodes representing states $s_t$ and edges representing actions $a_t$.
We consider the initial state $s_1$, which includes the task instruction $u$ and the initial observation $o_0$, as the root node.
A search trajectory starting from $s_1$ is formalized as a branch extending from the root node.
Each node records information such as the state content (action $a_t$ and corresponding observation $o_t$), the number of visit $N(s_t)$, and the expected future reward $V(s_t)$ starting from state $s_t$.
For each task instruction, we construct a search tree starting from the root node and expanding through repeating the following four stages for $\omega$ iterations:

\textbf{Selection} aims to identify the most promising node to be expanded in the next iteration.
Starting from the root node, it traverses the tree by selecting child nodes according to the Upper Confidence Bound (UCB) value until a leaf is reached:

\small\begin{equation*}
    s_{t}=\underset{s_j\in\mathrm{Children}(s_{t-1})}{\operatorname*{\operatorname*{\arg\max}}}\bigg(V(s_j)+c\cdot\sqrt{\frac{\log N(s_{t-1})}{1+N(s_j)}}\bigg),
\end{equation*}\normalsize

\textbf{Expansion} will be operated on the selected node $s_{t}$ if it is not a terminal state exceeding the maximum step or finishing reasoning.
The agent samples the next action $a_{t} \sim \pi(\cdot\mid s_{t})$ for $k$ times with temperature $\tau$ based on its policy.
Actions with identical action tokens are merged to lower the cost of repetitive search, resulting in $\hat{k}$ next states $\{s_{t+1}^i\}=\{(s_t,a_t,o_t)^i\}, i=1\dots\hat{k}$.

\textbf{Simulation} is used to estimate the initial value of the above expanded node $s_{t+1}$ by generating $n$ complete trajectories from it to get the outcome reward returned by the environment and averaging their outcome rewards.

To speed up the tree sear, we cache the rollout nodes for future expansion.

\textbf{Backpropagation} is conducted once the values of the expanded nodes are determined.
The value $V(s_{t+1}^i)$ is propagated back up the tree, updating each node's visit count $N$ and state value $V$:

\small\begin{align*}
    V(s_t) &\gets \frac{V(s_t) \cdot N(s_t) + \sum_{i=1}^{\hat{k}} V(s_{t+1}^{i})}{N(s_t) + \hat{k}}, \\
    N(s_t) &\gets N(s_t) + \hat{k}
\end{align*}\normalsize

\paragraph{Reward Model Training}
\label{sec:implicit}
For each task instruction in the held-in tasks i.e. Webshop, Alfworld, Sciworld, we construct a search tree and extract state values $V(s_t)$ to form the process reward model training dataset. 
To ensure the quality of the estimated value, we filter states whose visit count is smaller than threshold $\lambda$.
We train a language model with a value head by minimizing the Mean Squared Error (MSE) loss between the predicted value $\hat{V}(s_t)$ and the provided value $V(s_t)$:

\small\begin{equation}
    \mathcal{L}({\theta})=\frac{1}{N}\sum_{t=1}^{N}(\hat{V}(s_t) - V(s_t))^2
\end{equation}\normalsize

\subsubsection{Implicit Reward Modeling}

Implicit reward modeling typically defines process reward as advantage \cite{schulman2017proximal}, i.e. relative benefits of an action at a given state compared to alternatives.
It derives inherent process rewards from the model trained on outcome rewards, eliminating the overhead of process reward collection \cite{rafailov2024directpreferenceoptimizationlanguage, yuan2024freeprocessrewardsprocess}.
Specifically, the outcome reward is parameterized as the log-likelihood ratios of the policy and reference models, i.e. $r_{\theta}(s_T,a_T):=\beta \log\frac{\pi_{\theta}(s_T,a_T)}{\pi_{ref}(s_T,a_T)}$.
It is proved that the Q value $q_{\theta}^t(s_t,a_t)$ can be implicitly learned by $\theta$ (mathematical induction can be found in \cite{yuan2024freeprocessrewardsprocess}).
The process reward $r_{\theta}^t$ can be derived as follows:

\small\begin{equation}
    r_\theta^t:=q_{\theta}^t-q_{\theta}^{t-1}
    =\beta\log\frac{\pi_\theta(a_t\mid s_t)}{\pi_{\mathrm{ref}}(a_t\mid s_t)}
\label{eq:implicit}
\end{equation}\normalsize
where $\pi_\theta, \pi_{\mathrm{ref}}$ represent the policy and reference model parameter respectively.

\paragraph{Reward Model Training}
For each task instruction in the held-in tasks, we sample 16 complete trajectories $(s_T,a_T)$ with temperature $\tau$ to construct the process reward model training dataset.
We train a language model $\theta$ with the MSE loss to integrate the scalar reward (progress rate) provided by the environment, unlike \cite{yuan2024freeprocessrewardsprocess} using the cross-entropy loss for binary reward.

\subsubsection{LLM-as-a-judge}
\label{sec:llmasajudge}
In order to answer the question that can an LLM be used as the reward model to perform guidance without reward learning, we implement a training-free reward model following the paradigm of LLM-as-a-judge \cite{gu2025surveyllmasajudge}.
We prompt the LLM to act as a selector with instructions in Appendix~\ref{app:prompt}.

\subsection{Reward-Guided Search}
\label{sec:guide}

We boost the policy model at test time via search methods guided by our general reward model.

\noindent\textbf{Best-of-N} samples $N$ complete trajectories from the policy model and selects the final answer according to the output of the reward model.

\noindent\textbf{Beam Search} searches over the policy model's per-step generation in the following steps:

\begin{itemize}[noitemsep,topsep=0cm,leftmargin=0.4cm]
    \item Initial Sampling: Sample \( W_1 \times W_2 \) initial actions for the first step.
    \item Scoring: Evaluate the new states using the reward model.
    \item Filtering: Retain only the top \( W_1 \) highest-scoring states.
    \item Action Expansion: For each of the remaining states, sample \( W_2 \) actions for the next step, generating a total of \( W_1 \times W_2 \) new states.
    \item Iteration: Repeat steps 2–4 until all maintained states terminate.
\end{itemize}

\begin{table*}[htpb]
\centering
\setlength{\tabcolsep}{3pt}
 \small

\begin{tabular}{lllllllllll}
\toprule
   \multirow{2}{*}{\textbf{Method}}    & \multicolumn{1}{c}{\textbf{Web}} & \multicolumn{3}{c}{\textbf{Embodied}} &  \multicolumn{3}{c}{\textbf{Text Game}} & \multicolumn{2}{c}{\textbf{Tool}} & \multirow{2}{*}{\textbf{Overall}}\\ 
   \cmidrule(r){2-2} \cmidrule(r){3-5} \cmidrule(r){6-8} \cmidrule(r){9-10} &\textbf{Webshop}    &\textbf{Alfworld}    &\textbf{Sciworld}  &\textbf{Babyai}   &\textbf{Jericho}     &\textbf{Pddl}     &\textbf{Maze}    &\textbf{ToolQuery}     &\textbf{ToolOperation} \\ 
\midrule
gpt-4o
&57.7&79.9&76.9&64.1&34.0&69.8&76.0&61.8&37.6&65.9\\

Agent-FLAN
&61.3\textsuperscript{*}&79.7\textsuperscript{*}&10.9&35.3&10.1&25.5&44.0&45.7&26.8&47.1\\
AgentGym
&68.5\textsuperscript{*}&76.9\textsuperscript{*}&47.3\textsuperscript{*}&61.4\textsuperscript{*}&12.9&16.6&56.0\textsuperscript{*}&69.7\textsuperscript{*}&40.2\textsuperscript{*}&59.3\textsuperscript{*}\\
AgentGen
&53.9&47.6&13.9&39.4&10.8&36.4&44.0&57.6&25.1&42.0\\
AgentRefine
&-&63.8&42.6&50.4&32.3&37.8&-&-&-&-\\
ATLAS & 71.5\textsuperscript{*} & 84.5\textsuperscript{*} & 42.0\textsuperscript{*} & 81.0\textsuperscript{*} & 18.2 & 15.8 & 48.0\textsuperscript{*} & 73.3\textsuperscript{*} & 69.7\textsuperscript{*}& 64.9\textsuperscript{*}\\
Greedy Search
&57.8&51.1&48.5&52.1&22.5&37.7&52.0&76.1&41.6& 52.7\\
\midrule
\multicolumn{11}{c}{\textit{Best-of-5}} \\
\midrule
Explicit RM 
&62.4&67.7&50.1&70.6&\textbf{30.0}&33.3&\textbf{80.0} &82.1&\textbf{43.9}&61.5\\

Implicit RM 
&60.5&61.8&35.4&58.2&23.3&26.0&68.0&81.2&38.8&54.7\\
LLM-as-a-judge 
&55.6&59.0&29.3&58.3&20.3&22.9&72.0&\textbf{83.1}&41.9&52.1\\

\midrule
\multicolumn{11}{c}{\textit{Beam Search ($W_1=5, W_2=5$)}} \\
\midrule

Explicit RM

&\textbf{64.4}&\textbf{72.4}&\textbf{51.7}&\textbf{71.2}&29.1&\textbf{41.4}&72.0&79.3&40.6&\textbf{63.3}\\
\bottomrule
\end{tabular}
\caption{Performance comparison with general agents.\textsuperscript{*} indicates the task is seen during policy training and treated as held-in evaluation. Overall performance is averaged across tasks, weighted by test set sizes. LLaMA3-8B-Instruct is used for all methods.} 

\label{tab:main}
\end{table*}

\section{Experiments}

\subsection{Baselines}
Apart from greedy search, we compare our method with task-specific agents and general agents.
Task-specific agents include SPIN \cite{chen2024self}, NAT \cite{wang2024learning}, ETO \cite{song2024trial}, StepAgent \cite{deng2024novice}, QLASS \cite{lin2025qlassboostinglanguageagent}, Agent-R \cite{yuan2025agent}, MPO \cite{xiong2025mpo}, and GLIDER \cite{hu2025divideconquergroundingllms}.
General agents include Agent-FLAN \cite{chen2024agent}, AgentGym \cite{xi2024agentgym}, AgentGen \cite{hu2024agentgen}, AgentRefine \cite{fu2025agentrefineenhancingagentgeneralization}, ATLAS \cite{chen2025atlas}.
We also compare our method with close-sourced agent based on gpt-4o for reference.
More details can be found in Appendix~\ref{app:baseline}.

\subsection{Experimental Settings}
\paragraph{Datasets}
We adopt the three agent tasks from ETO \cite{song2024trial} as our held-in tasks, which refers to the environments that are seen during training: Webshop for web navigation, Alfworld for embodied house holding, and Sciworld for embodied science experiments.
We adopt agent tasks from AgentBoard \cite{ma2024agentboard} and AgentGym \cite{xi2024agentgym} as held-out tasks, which refers to the environments unseen during training: Babyai for embodied planning, Jericho and Pddl and Maze for text-based game, ToolQuery (including Weather, Movie, Academia) and ToolOperation (including TODO and Sheet) for tool using.
Note that there are two sources of Alfworld and Sciworld. In order to align with the setting of previous works, we use the former to train the RM and evaluate in Section~\ref{sec:specific_agent}, while the latter is used for evaluation in Section~\ref{sec:general_agent}.
Details can be found in Appendix~\ref{app:statistics}.

\paragraph{Metrics}
Maze and Alfworld(ETO) provide \textbf{Success Rate} indicating whether a task is successfully completed.
Others provide \textbf{Progress Rate}, a scalar measuring the completion percentage.
We use the average reward as the metric for each task.

\paragraph{Implementation Details}
We adopt the LLaMA3-8B-Instruct series model as our policy model and reward model.
More details can be found in Appendix~\ref{sec:details}.
We split 1/4 of the expert trajectories for SFT, i.e. 1938, 830, 370 for Webshop, Alfworld, Sciworld.
The remaining 3/4 instructions are used to train reward model without expert annotation.

\begin{table}[htpb]
\centering
\setlength{\tabcolsep}{6pt}

\small
\begin{tabular}{lccc}
\toprule

    \textbf{Method} & \textbf{Webshop} & \textbf{Alfworld\textdagger} & \textbf{Sciworld\textdagger} \\
\midrule

gpt-4o &57.7&66.4&66.6\\
SPIN
&65.4&71.9&60.3\\

NAT & 63.2 & 68.3 & 55.6 \\
ETO
&65.7&73.4&62.5\\

StepAgent&67.6&76.1&64.1\\

QLASS & 70.3 & 82.8 & 66.4 \\
Agent-R & 63.9 & - & 70.2\\
MPO & - &79.1 &80.8\\
GLIDER &- & 75.4 & 68.3\\
Greedy Search & 61.4& 71.6& 66.6\\
\midrule
\multicolumn{4}{c}{\textit{Best-of-5}} \\
\midrule
Explicit RM&71.0&94.8&76.1\\
Implicit RM &66.4&94.8&70.6 \\
LLM-as-a-judge &60.5&64.9&62.3 \\
\midrule
\multicolumn{4}{c}{\textit{Beam Search ($W_1=5, W_2=5$)}}\\
\midrule
Explicit RM&\textbf{75.3}&\textbf{96.3}&\textbf{82.6}\\
\bottomrule
\end{tabular}
\caption{Comparison with task-specific agents. \textdagger means the sources of Alfworld and Sciworld differ from those in Table~\ref{tab:general} thus incomparable, detailed in Appendix~\ref{app:statistics}. LLaMA3-8B-Instruct is used for all methods.} 
\label{tab:specific}
\end{table}

\begin{table*}[htbp]
\centering
\setlength{\tabcolsep}{2pt}
\small
\begin{tabular}{lllllllllllllllrr}
\toprule

   \multirow{2}{*}{\textbf{Method}}    & \multicolumn{2}{c}{\textbf{Original}} &  \multicolumn{2}{c}{\textbf{Rule 1}}    & \multicolumn{2}{c}{\textbf{Rule 2}}    & \multicolumn{2}{c}{\textbf{Rule 3}}    & \multicolumn{2}{c}{\textbf{Rule 4}}    & \multicolumn{2}{c}{\textbf{Rule 5}}    & \multicolumn{2}{c}{\textbf{Average($\uparrow$)}}    & \multicolumn{2}{c}{\textbf{Std($\downarrow$)}} \\ 
   \cmidrule(r){2-3} \cmidrule(r){4-5} \cmidrule(r){6-7} \cmidrule(r){8-9} \cmidrule(r){10-11} \cmidrule(r){12-13} \cmidrule(r){14-15} \cmidrule(r){16-17}  &\textbf{Succ.}    &\textbf{Prog.}    &\textbf{Succ.}    &\textbf{Prog.}  &\textbf{Succ.}    &\textbf{Prog.}    &\textbf{Succ.}    &\textbf{Prog.}    &\textbf{Succ.}    &\textbf{Prog.}  &\textbf{Succ.}    &\textbf{Prog.}   &\textbf{Succ.}    &\textbf{Prog.}   &\textbf{Succ.}    &\textbf{Prog.}\\ 
\hline
AgentGym
&61.9\textsuperscript{*}&76.9\textsuperscript{*}&29.1&59.2&49.2&65.3&32.8&53.9&38.8&48.2&5.9&28.7&36.3&55.4&20.0&16.7\\
Agent-FLAN
&67.2\textsuperscript{*}&79.7\textsuperscript{*}&21.6&58.8&51.4&71.3&27.6&53.5&52.2&67.9&1.5&19.7&36.9&58.5&22.0&22.5\\
AgentRefine
&44.8&63.8&50.0&66.5&51.5&66.7&54.5&70.0&45.5&60.6&44.8&63.8&48.5&65.2&4.1&3.2\\
Ours

&54.5&67.7&54.5&68.6&53.0&70.2&48.5&63.6&49.3&63.9&54.5&67.7&\textbf{52.4}&\textbf{66.9}&\textbf{2.7}&\textbf{2.6}\\
\bottomrule
\end{tabular}
\caption{Performance of Alfworld under different perturbation rules. Succ./Prog. denote Success/Progress Rate respectively. $^*$ indicates the task is seen during training and treated as held-in evaluation.} 
\label{tab:robust}
\end{table*}

\subsection{Results}
\subsubsection{Comparison with General Agents}
\label{sec:general_agent}

In this setting, we compare our method with methods that aim to train a single unified agent for various tasks.
To make a fair comparison, we use the original non-finetuned model as the policy model since fine-tuning leads to performance degradation on held-out tasks, and guide its generation with our AgentRM.
From Table~\ref{tab:main} we can observe that: (1) Existing general agents exhibit severe overfitting in held-in tasks, as their overall performance fail to substantially surpass those of the greedy search baseline.
While AgentGym and ATLAS achieves a high score, it is primarily because most of the task environments are seen during training. This advantage, however, is offset by its notably weak performance on held-out tasks i.e. Jericho and Pddl.
(2) Three types of AgentRM bring varying degrees of improvement over the baseline. Among them, Explicit RM proves to be the most effective, enhancing the greedy search baseline by $8.8$ on average.
(3) On the Babyai task, which shares similarities with the held-in tasks Alfworld and Sciworld, the explicit RM exhibits significant positive transfer. Conversely, we observe that a policy model trained on Sciworld but not on Babyai tends to overfit to the action space of Sciworld, leading to negative transfer.
(4) Best-of-5 with LLM-as-a-judge shows a $0.6$ decline on overall performance compared to greedy search, suggesting that LLM of 8B cannot be used to guide the inference effectively without reward learning.
Among all tasks, it performs relatively better on tool-related tasks, suggesting that LLM-as-a-judge is more effective on tasks with less complexity and smaller search space, while being less effective on complex tasks.

\subsubsection{Comparison with Task-specific Agents}
\label{sec:specific_agent}

In this setting, we compare our method with methods that aim to train a specialized agent for each task.
Instead of training task-specific policy models, we find a single policy model simultaneously trained on three tasks capable of mastering each task without compromising performance on any.
Out of the same reason, we use the general RM same as Section~\ref{sec:general_agent} without task-specific fine-tuning.
From the results in Table~\ref{tab:specific}, Best-of-5 with Explicit RM enhances the policy model by 9.6, 23.2 and 9.5 on three held-in tasks respectively.
It outperforms top specialized agents including Agent-R and QLASS across all tasks, showing potential in more practical scenarios where an agent is required to be proficient in more than one task \cite{acikgoz2025singlemodelmastermultiturn}.
Further improvements can be achieved through beam search.

\section{Analysis}
In the following analysis, unless otherwise stated, we report the results of explicit RM with Best-of-5 inference, as it outperforms the other two reward models notably.

\subsection{Robustness against Perturbation}

To test the extent of overfitting on the held-in tasks, we perform 5 types of perturbations on the held-in task.
Specifically, we perturb available actions in the task instruction of Alfworld, which belongs to the held-in tasks for AgentGym and Agent-FLAN.
See Appendix~\ref{sec:perturb} for details of perturbation rules.

From Table~\ref{tab:robust} we can see that, simple data perturbation leads to a significant performance drop on the held-in task. 
In terms of the average score, AgentGym's success rate decreases by 25.6, whereas Agent-FLAN shows a more significant performance drop of 30.3.
This suggests that they might simply be memorizing the correlations between instructions/observations and corresponding actions from the training data, rather than learning to respond to the given instructions and observations.
Our method achieves the highest average score with the lowest standard deviation, 
indicating that it develops the ability to make informed decisions, rather than memorizing patterns.

\begin{figure}[b]
    \centering
    \includegraphics[width=0.95\linewidth]{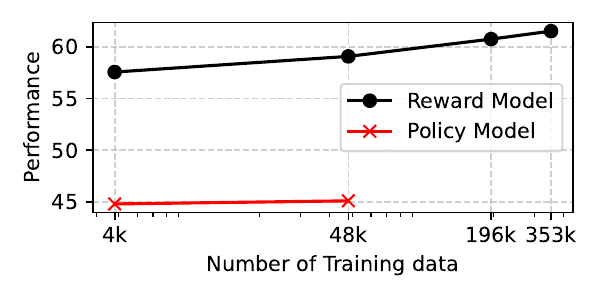}
    \caption{Scaling trend of training data.}
    \label{fig:data_scaling}
\end{figure}

\subsection{Scaling Trend of Training Data}
\label{sec:data_scaling}
We analyze the relationship between the training data size of the reward model and overall performance, with the results shown in Figure~\ref{fig:data_scaling}.
The results demonstrate that even a relatively small dataset of 4k states is able to elicit significant reward modeling capabilities (57.6) for agent tasks, compared to the prompt-based training-free LLM-as-a-judge (52.1).
This underscores the effectiveness of our approach in data-constrained scenarios.
As the volume of training data increases, the performance exhibits a persistent log-linear growth without showing signs of saturation. 
The observed trend leaves room for continued performance optimization with expanded datasets.

Besides, we compare standalone policy model training with standalone reward model training using equivalent amounts of training data.
Since the volume of trajectory-level policy model data is much smaller than state-level reward model data, we try two ways to align their scales:
(1) Downsampling the amount of the reward model training data to $\sim$4k.
(2) Upsampling the SFT training data to $\sim$48k by collecting successful trajectories from the constructed MCTS trees.
Results in Figure~\ref{fig:data_scaling} demonstrate that standalone reward model training consistently outperforms standalone policy model training when trained on equivalent amounts of data, and is more scalable.

\subsection{Generalization of Task-specific RM}
\begin{figure}[t]
    \centering
    \includegraphics[width=\linewidth]{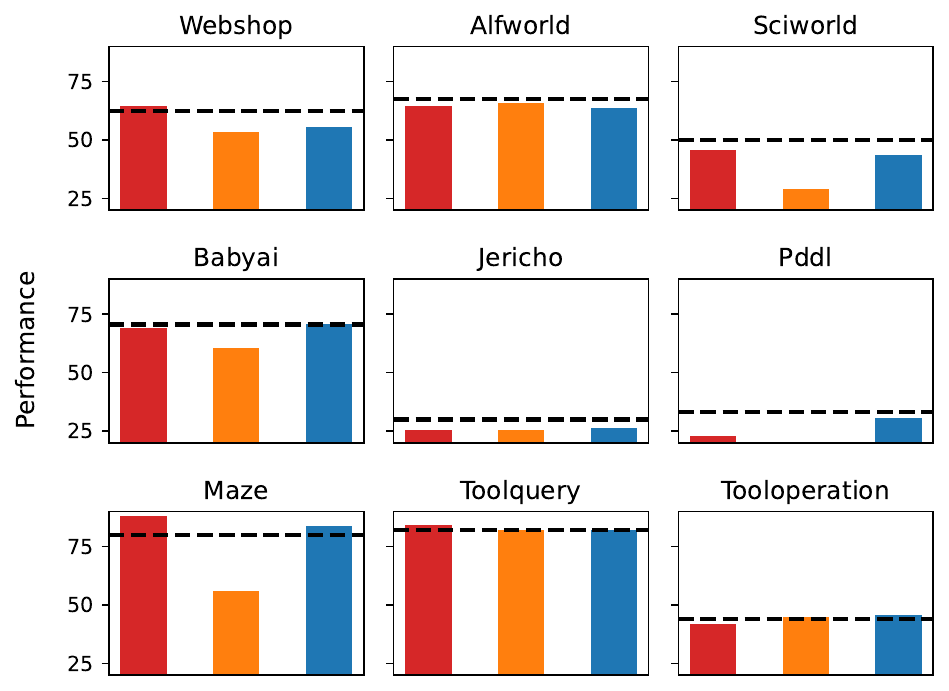}
    \caption{Performance of task-specific RM on 9 tasks. The red/orange/blue bar denotes RM trained on Webshop/Alfworld/Sciworld respectively. The dashed line denotes the performance of the general RM.}
    \label{fig:task_contribution}
\end{figure}

We examine the generalization of task-specific RM trained on each held-in task (Figure~\ref{fig:task_contribution}).
The results reveal that, for most tasks, the general RM (dashed line) outperforms task-specific RMs, verifying the importance of task diversity in enhancing RM generalization.
Besides, the task-specific RM trained on Alfworld exhibits comparatively weaker performance, which may be attributed to the use of success rate rather than the progress rate, which is a denser signal, as the outcome supervision when constructing RM training data.

\subsection{Plug-and-play for Other Policy Models}
\label{sec:policy}
It is commonly thought that broad training data coverage is a requirement to ensure adaptability to different policy distribution. 
Surprisingly, we find that our RM, which is trained only on states sampled by the LLaMA-3-8B policy model, can be effectively applied to states sampled by other LLM agents.
Specifically, we directly utilize our RM to supervise different policy models including LLaMA-3.2-1B, LLaMA-3.2-3B, AgentGen (LLaMA-3-8B finetuned on synthetic data), Qwen2.5-14B-Instruct, and LLaMA-3-70B.
From Table~\ref{tab:otherpolicy} we can see that our RM adapts well to different policy models and consistently improves the overall performance.
Specifically, it improves the LLaMA-3-70B-based agent by 12.6 and AgentGen by 5.9, demonstrating more pronounced advantages for models with greater scale and potential.
These encouraging results indicate that the trial-and-error task experience derived from a weaker yet more efficient agent can enhance the performance of stronger and more costly agents, facilitating weak-to-strong generalization \cite{yang2024weak}.
\begin{table*}[t]
\centering
\setlength{\tabcolsep}{2pt}
\small
\begin{tabular}{lccccccccccc}
\toprule
   \textbf{Method}    
   & \textbf{Webshop}    &\textbf{Alfworld}  &\textbf{Sciworld}  
   &\textbf{Babyai}   &\textbf{Jericho}     &\textbf{Pddl}     &\textbf{Maze}    &\textbf{Toolquery}     &\textbf{Tooloperation} & \textbf{Overall} \\ 
\midrule
\multicolumn{10}{c}{\textit{LLaMA-3-70B}} \\
\midrule

Greedy Search (w/o RM) &63.4&63.4&51.1& 62.6&31.7&64.1&76.0& 81.9&44.9&62.4\\
BestofN@5 (w/ RM) &69.5&86.9&78.8&72.0&43.0&67.9&96.0 & 84.6 &45.9&74.9\\
$\Delta$ &6.1 &23.5&27.7&9.4 & 11.3 & 3.9 & 20.0 & 2.7 & 1.0&12.6 \\
\midrule
\multicolumn{10}{c}{\textit{Qwen2.5-14B-Instruct}} \\ 
\midrule
Greedy Search (w/o RM) &64.3&29.7&49.1&49.0&29.3&56.6&60.0&80.0&25.1	&52.1\\
BestofN@5 (w/ RM) & 68.2&45.3&61.6&67.9&35.9&66.1&72.0&48.3&25.3	&59.3 \\
$\Delta$ & 3.9&15.6&12.5&18.9&6.6&9.5&12&-31.7&0.2&7.2\\
\midrule
\multicolumn{10}{c}{\textit{AgentGen (LLaMA-3-8B fine-tuned on synthetic data)}} \\
\midrule
Greedy Search (w/o RM) &53.9&29.1&13.9&39.4&10.8&36.4&44.0&57.6&25.1&38.6\\
BestofN@5 (w/ RM) &58.7&45.0&10.6&44.6 &14.7&42.9&44.0&62.8&30.2&44.4 \\
$\Delta$ & 4.7 &15.9&-3.3&5.1 & 4.0 & 6.5 & 0.0 & 5.2 & 5.1 & 5.9\\
\midrule
\multicolumn{10}{c}{\textit{LLaMA-3.2-3B}} \\ 
\midrule
Greedy Search (w/o RM) & 46.8&33.4&28.1&49.9&18.6&12.3&20.0&	53.6&8.2&36.7\\
BestofN@5 (w/ RM) & 57.7&34.4&28.1&59.8&20.9&12.1&24&66.4&11.6&	43.5\\
$\Delta$ & 10.9&1.0&0.0 &9.9&2.3&-0.2&4.0&	12.8&3.4&6.8\\
\midrule
\multicolumn{10}{c}{\textit{LLaMA-3.2-1B}} \\ 
\midrule
Greedy Search (w/o RM) &22.1&20.8&4.0&	26.1&2.0&	3.8&	4.0&	15.2&0.0&	16.3 \\
BestofN@5 (w/ RM) & 33.2&18.3&0.2&30.4&2.8&2.8&16&12.3&0.0&	19.2\\
$\Delta$ & 11.1&-2.5&-3.8&4.3&0.8&-1.0&	12.0&-2.9&0.0&2.9 \\
\bottomrule
\end{tabular}
\caption{Enhancement of our AgentRM to other policy models. Generally, the models with greater scale and potential achieve more pronounced improvement.}
\label{tab:otherpolicy}
\end{table*}

\subsection{State Representation of Reward Modeling}

As stated in Section~\ref{sec:rm}, the input of our RM consists of thought tokens, action tokens, and observation tokens (except those of the last action).
This section examines their respective contributions to the overall performance.
Results are shown in Table~\ref{tab:ablation}.
Explicit RM w/ last\_observation means adding the observation of the last action to the state representation during both training and inference.
It can be seen that the determination of state rewards for different tasks has varying degrees of reliance on the outcomes of actions.
Overall, augmenting the action with its outcome does not bring significant improvement, suggesting that the RM might possess the ability to infer the consequence autonomously. 
Results w/o observation and w/o thought show that the individual removal of thought and observation has a negligible impact on the modeling.
Results w/o thought \& observation show that removing them simultaneously results in a drop of 3.2 points, indicating that thought and observation tokens provide complementary information to each other.
In conclusion, the modeling primarily relies on action tokens. 
Utilizing only action tokens for modeling does not significantly impact the effectiveness and can accelerate the training and inference of the reward model, promoting scalability.

\begin{table*}[htpb]
\centering
\setlength{\tabcolsep}{2pt}
\small
\begin{tabular}{lcccccccccc}
\toprule
\textbf{Method}    
& \textbf{Webshop}    &\textbf{Alfworld}    &\textbf{Sciworld}  
&\textbf{Babyai}   &\textbf{Jericho}     &\textbf{Pddl}     &\textbf{Maze}    &\textbf{Toolquery}     &\textbf{Tooloperation} & \textbf{Overall} \\ 
\midrule
Explicit RM &62.4&67.7&50.1&70.6&30.0&33.3&80.0 &82.1&43.9&61.5\\
w/ last\_observation &62.4& 66.7&52.2&73.3&30.6&32.2&80.0&82.2&43.9&62.0\\
w/o observation &63.7&68.0&43.4&71.3&23.4&31.0&88.0&83.0&43.9&61.2\\
w/o thought &62.0&66.5&48.7& 71.1&32.1&30.2&84.0&82.9&44.9&61.1\\
w/o thought \& observation &62.4&66.0&45.7&69.1&22.1&25.4&44.0&83.2&39.4&58.3 \\
\bottomrule
\end{tabular}
\caption{Ablation on state representation of Explicit RM.}
\label{tab:ablation}
\end{table*}

\subsection{Scaling Trend of Test-time Search}
\begin{figure}[t]
    \centering
    \includegraphics[width=0.9\columnwidth]{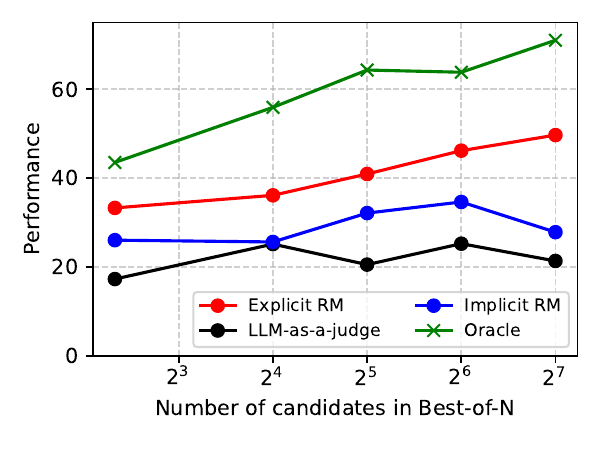}
    \caption{Scaling trend of Best-of-N.}
    \vspace{-1.5em}
    \label{fig:bestofn}
\end{figure}

We select Pddl task to explore the potential gains from further increasing the number of candidates in the Best-of-N sampling using different reward modelings.
The oracle result is obtained by selecting the best candidate based on the ground-truth label, which is not feasible in practice. We report it as an upper bound of performance.
As shown in Figure~\ref{fig:bestofn}, explicit RM yields consistent performance gains as the test-time compute increases.
When the number of candidates increases to a certain extent, the implicit RM may become confused by the excessive number of candidates, leading to a degradation in performance.
The effectiveness of using LLM-as-a-judge for scaling is limited. 
One reason is that as N increases, a growing number of tokens exceeding the maximum token limit of the model will be truncated.
The findings indicate that additional research is necessary to establish robust test-time scaling laws with Implicit RM and LLM-as-a-judge, which we leave for future work.


\subsection{Preservation of General Reasoning Capabilities}
The relationship between agent tasks and general reasoning tasks remains unclear.
In this section, we explore the impact of our RM, merely trained on agent tasks, on the general reasoning tasks.
We directly apply our RM on several general reasoning benchmarks including GSM8k \cite{cobbe2021gsm8k}, MATH \cite{hendrycks2021measuring} and codecontests \cite{li2022competition}.
We prompt the policy model to solve mathematical problems using a Python interpreter.
Table~\ref{tab:general} shows that, our RM trained on agent tasks has a negligible impact on general reasoning tasks, indicating the RM has acquired reasoning abilities common to general reasoning tasks, rather than merely fitting the patterns of agent tasks.

\begin{table}[t]
\centering
\setlength{\tabcolsep}{2pt}
\small
\begin{tabular}{lccccc} 
\toprule
   \textbf{Method}    
   &\textbf{GSM8k}   &\textbf{MATH500} & \textbf{Codecontests} \\
\midrule
Greedy Search & 81.1 & 48.4&13.3\\
BestofN@5 &79.1&49.2&13.9\\
\bottomrule
\end{tabular}
\caption{Performance on general reasoning tasks.} 
\label{tab:general}
\end{table}

\section{Related Work}
\subsection{LLM-based Agent}
Language agents have shown initial success in handling complex interactive tasks.
Early works focus on building frameworks around prompt-based learning \cite{yao2022react, shinn2024reflexion}.
Recently, great efforts have been made to enhance the agent capability of open-sourced LLMs via finetuning \cite{chen2023fireact, yin2024agent}.
\citet{qin2023toolllm, deng2024mind2web} imitate trajectories from expert agents (e.g., GPT-4 \cite{achiam2023gpt}) for specialized ability such as tool-using or web navigation.
Beyond imitation, self-improvement emerges as a promising solution to enhance performance without extensive expert annotation \cite{huang-etal-2023-large}. 
Most works finetune models on self-generated trajectories following the self-training paradigm \cite{wang2024learning, chen2024self, song2024trial, xiong2024watch, ye2024rationaldecisionmakingagentinternalized}.
Lately, increasing attention has been devoted to self-improvement via test-time computation, e.g., generating multiple candidates and selecting the optimal one using techniques like reward models \cite{wang2024q, zhai2024enhancingdecisionmakingllmagents, lin2025qlassboostinglanguageagent, chen2025scaling}.
We provide a comparison between their approaches and our method in Section~\ref{sec:rmforllm}.

While effective for tasks seen during training, the above methods inherently compromise the agent's generalization capabilities for unseen tasks.
To enhance agent generalizability, existing works integrate more diverse agent tasks for multi-task training either by human-crafted \cite{zeng2023agenttuning, chen2024agent, xi2024agentgym, zhang2024agentohana} or by LLM-sythesized \cite{hu2024agentgen, fu2025agentrefineenhancingagentgeneralization}.
Although they alleviate overfitting to some extent, it can be observed in Table~\ref{tab:main} that their performance on respective held-out tasks is either similar or inferior to that of the original backbone model.
We are the first to propose a generalizable reward model and enhance the agent generalizability from the aspect of test-time search.
Also, our method is orthogonal to theirs and can be applied to enhance their performance seamlessly, as shown in Section~\ref{sec:policy}.

\subsection{Reward Modeling for LLM}
\label{sec:rmforllm}
Recent advancements in reward modeling for LLMs mainly focus on general reasoning tasks such as maths and code \cite{uesato2022solving,lightman2023let,wang2023math, zhang2024rest}.
Different from those tasks, agent tasks typically possess a larger search space due to long-chain reasoning and environment dynamics.
Data scarcity is also a challenge pronounced in agent tasks \cite{ma2024agentboard}, making it impractical to develop task-specific reward models.
Relevant works on agent tasks \cite{wang2024q, zhai2024enhancingdecisionmakingllmagents, putta2024agent, lin2025qlassboostinglanguageagent} focus on training task-specific process reward models by Tree Search based methods.
We are the first to investigate the feasibility of a generalizable reward model, promoting the usage of reward models in agent tasks.
Besides, we investigate two additional reward modelings and validate them on six additional complex agent tasks with larger search space.

\section{Conclusion}
We introduce AgentRM, a generalizable reward model, which enhances the performance of language agents via test-time search.
We comprehensively investigate three reward modelings, i.e. explicit reward modeling, implicit reward modeling and LLM-as-a-judge.
Among them, explicit reward modeling achieves the best performance.
Extensive experiments on nine agent tasks show the effectiveness of our method in both specializability and generalizability.
Moreover, it demonstrates weak-to-strong generalization, yielding greater improvement on more powerful policy models.
We hope this work shed light on generalization and test-time self-improvement of agents.

\section*{Limitations}
We conclude the limitations of this work as follows:
\begin{itemize}
    \item Due to the manual effort required to implement additional agent interactive environments, we only include three agent tasks as held-in tasks. According to the scaling trend of training data in Section~\ref{sec:data_scaling}, incorporating more tasks could further enhance the performance.
    \item We do not explore the potential of equipping our policy model with prompt engineering designed for agent such as Reflexion \cite{shinn2024reflexion}.
    \item In this study, we focus on applying outcome and process supervision to the reward model. While fine-tuning the policy model using reinforcement learning (RL) would be a natural extension, we defer this to future work. Instead, we concentrate on the contributed dataset and demonstrate that process supervision alone achieves performance comparable to RL-based methods \cite{feng2025group}.
\end{itemize} 

\section*{Acknowledgments}
This work was supported in part by the National Natural Science Foundation of China (Grant No. 62376273), the Postdoctoral Fellowship Program of CPSF (Grant No. GZB20230343 and Grant No. GZC20240831) and the China Postdoctoral Science Foundation (Grant No. 2023M741945).

\bibliography{main}

\appendix

\section{Perturbation Details}
\label{sec:perturb}
We modify the available actions in Alfworld to ensure that the changes consist of different tokens (or token order) while conveying the same semantic information.
We revise the environment and the examples in the prompt accordingly.
\begin{itemize}
    \item Perturbation 1: change clean $\{obj\}$ with $\{recep\}$, cool $\{obj\}$ with $\{recep\}$, heat $\{obj\}$ with $\{recep\}$ to clean $\{obj\}$ using $\{recep\}$, cool $\{obj\}$ using $\{recep\}$, heat $\{obj\}$ using $\{recep\}$ in the instruction
    \item Perturbation 2: change go to $\{recep\}$ to move to $\{recep\}$ in the instruction
    \item Perturbation 3: change take $\{obj\}$ from $\{recep\}$ to from $\{recep\}$ take $\{obj\}$ in the instruction
    \item Perturbation 4: delete all space between item name and item number in the instruction
    \item Perturbation 5: remove all alfworld data in the training set and retrain the model
\end{itemize}

\section{Implementation Details}
\label{sec:details}
Hyperparameters are listed in Table~\ref{tab:hyper}.
The SFT data is obtained by randomly selecting 1/4 expert trajectories from the training set.
Note that the data is formatted in ReAct-style \cite{yao2022react}, and  $a$ in Section~\ref{sec:bc} denotes the complete ReAct-style response (containing both thought and action tokens) generated by $\pi$.
The remaining 3/4 of the data is reserved for constructing RM training data.
In the explicit reward data construction stage, we set the iteration number $\omega$ as 40, the exploration constant $c$ in UCB as 0.5, the filtering threshold $\lambda$ as 3, the number of the rollout in simulation $n$ as 1, the rollout policy as greedy, the expansion width $k$ as 5.
We leverage the AdamW optimizer.
All experiments are carried out on 8 NVIDIA A100 80G GPUs.
We use vLLM \cite{kwon2023efficient} to implement both the policy model and reward model during inference.

\begin{table}[htpb]
\centering
\small
\begin{tabular}{cccc}
\toprule
   Stage  &SFT    &\makecell{Explicit RM\\Training}   &\makecell{Implicit RM\\Training}\\ 
\midrule
Learning Rate&2e-5&1e-5&5e-7\\[0.5em]
\makecell{Cosine Scheduler\\Warm Up}&0.1 &0.03& 5e-7\\[1em]
Batch Size&64 &96 &64 \\
Weight Decay &0.0&0.0&0.0 \\
Epoch &3 &2 &1\\
$\beta$&-&-&0.05\\
\bottomrule
\end{tabular}
\caption{Training hyper-parameters of different stages.} 
\label{tab:hyper}
\end{table}

\section{Baselines}
\label{app:baseline}
\subsubsection{General Agents}
\textbf{Agent-FLAN} \cite{chen2024agent} is an improvement of AgentTunning focusing on training "thought" in ReAct.
\textbf{AgentGym} \cite{xi2024agentgym} enables the model to continuously learn new tasks and treating all tasks as held-in via SFT and DPO.
\textbf{AgentGen} \cite{hu2024agentgen} uses LIMA to synthesize diversified agent-tuning data.
\textbf{AgentRefine} \cite{fu2025agentrefineenhancingagentgeneralization} 
proposes an environment synthesis method and distills the self-refinement ability from advanced proprietary models such as deepseek and gpt-4o via SFT.
For a fair comparison, all general agents receive the task instruction and one successful trajectory as input and respond in ReAct-style.
For a fair comparison, we reproduce Agent-FLAN, AgentGym and AgentGen based on LLaMA-3-8B-Instruct.
Agent-FLAN includes Alfworld in its training set. 
AgentGym includes Alfworld, BabyAI, and SciWorld in its training set. These datasets will be seen as held-in test tasks for the corresponding method.
Since AgentRefine has not open sourced, we only report the results on five tasks in \cite{fu2025agentrefineenhancingagentgeneralization} with LLaMA-3-8B-Instruct backbone.
\subsubsection{Task-specific Agents}
\textbf{SPIN} \cite{chen2024self} 
augments the expert trajectory dataset with the agent's successful trajectories.
\textbf{NAT} \cite{wang2024learning} and \textbf{ETO} \cite{song2024trial} incorporate failed trajectories into the training process, allowing the agent to learn from its failure experiences.
\textbf{StepAgent} \cite{deng2024novice} utilizes step-wise reward to optimize the agent’s reinforcement learning process.
\textbf{QLASS} \cite{lin2025qlassboostinglanguageagent} guides stepwise search with trained task-specific Q-value models.
\textbf{Agent-R} \cite{yuan2025agent} leverages MCTS to construct training samples that recover correct trajectories from erroneous ones.
Results of SPIN, NAT, ETO, StepAgent are taken from \cite{deng2024novice} with LLaMA-3-8B-Instruct backbone.
Since QLASS has not open sourced, we report the results in \cite{lin2025qlassboostinglanguageagent} with LLaMA-2-chat backbone.

\section{Task Statistics}
\label{app:statistics}
Table ~\ref{tab:statistics} presents the statistics of both held-in and held-out tasks.
We adopt the three agent tasks from ETO \cite{song2024trial} as our held-in tasks: Webshop for web navigation, Alfworld for embodied house holding, and Sciworld for embodied science experiments.
We adopt agent tasks from AgentBoard \cite{ma2024agentboard} and AgentGym \cite{xi2024agentgym} as held-out tasks: \textbf{Alfworld}, \textbf{Sciworld}, \textbf{Babyai} for embodied house holding, \textbf{Jericho} and \textbf{Pddl} and \textbf{Maze} for text game, \textbf{ToolQuery} and \textbf{ToolOperation} for tool using.
Note that there are two sources of Alfworld and Sciworld, i.e. ETO \cite{song2024trial} and AgentBoard \cite{ma2024agentboard}.
The reward model training data is collected through interactions with the ETO environment since it provides training set along with expert trajectories.
Evaluation in Section~\ref{sec:general_agent} / Section~\ref{sec:specific_agent} are conducted on Alfworld and Sciworld implemented by AgentBoard / ETO  respectively to align with previous works.
They have slight differences in \textbf{action space}, \textbf{test set number} and \textbf{metric}.

\begin{table*}[htpb]
\centering
\setlength{\tabcolsep}{2pt}
\scalebox{0.8}{
\begin{tabular}{cccccccccc}
\toprule
   task   &Webshop    &Alfworld   &Sciworld   &Babyai&Jericho    & PDDL      &Maze
  &Toolquery    &Tooloperation\\ 
\midrule
\# Train
&10426&3321&1483&-&-&-&-&-&-\\
\# SFT & 1938 & 830 &370&-&-&-&-&-&- \\
\# RM Training &8488& 2491 & 1113&-&-&-&-&-&- \\
\# Test
&200&134/134&211/90&112&20&60&25&60&40\\
Reward Type
&Prog.&Succ./Prog.&Prog./Prog.&Prog.&Prog.&Prog.&Prog.&Prog.&Prog.\\
Avg. Turn &3 & 6 & 15 & 10 & 20&20&4.3 & 5&6\\
Max. Turn
&10&20/30&$[15, 120]$/30&30&30&30&30&30&30\\
Action Space & 2 & 10/13 & 19/21 & 8&150&8&4 &15&16 \\
\bottomrule
\end{tabular}
}
\caption{Statistics of held-in and held-out tasks. Prog./Succ. denotes Progress/Success Rate. For the test set of Alfworld and Sciworld, we report the size of ETO (left) and AgentBoard (right).} 
\label{tab:statistics}
\end{table*}

\subsection{LLM-as-a-judge prompt}
\label{app:prompt}
We do not prompt the LLM to output a discrete score for each trajectory since the score might be identical thus insufficient to select the best answer from a set of candidates (e.g., Best-of-N).
Instead, we prompt the LLM as follows:
\begin{lstlisting}[language=Python]
You are trajectory reward model, an expert in defining which trajectory is better and closer to solving the task. Here is the task description:
*******************************
task description: {task_description}
task goal: {task_goal}
*******************************
Here are several candidates. They are all trying to solve the task. Their trajectories are as follows.
*******************************
CANDIDATE1:
{candidate_1}
*******************************
CANDIDATE2:
{candidate_2}
*******************************
CANIDATE3:
{candidate_3}
*******************************
CANIDATE4:
{candidate_4}
*******************************
CANIDATE5:
{candidate_5}
*******************************
\end{lstlisting}
We force the LLM to call the following function to give the answer:
\begin{lstlisting}[language=Python]
[{
    "type": "function",
    "function": {
"name": "choose_preferred_answer",
"description": "Choose the preferred answer for the task within all given answers.",
"parameters": {
"type": "object",
"properties": {
"preference": {
"type": "number",
"enum": [1, 2, 3, 4, 5],
"description": "The index of the preferred answer in all given answers (ranging from 1 to 5)."
},
},
}
    }
}]
\end{lstlisting}
\begin{table}[!t]
\setlength{\tabcolsep}{1pt}
\centering
\scalebox{0.85}{
\small
\begin{tabular}{lccccccc} 
\toprule
   Method    
   &Babyai   &Jericho     &Pddl     &Maze    &Toolquery     &Tooloperation \\ 
\midrule
LLM-as-a-judge & 65.7 & 46.0&70.7&65.8&81.4&42.1\\
Explicit RM & 77.0 & 64.9 & 65.4&94.7&72.9&57.9\\
\bottomrule
\end{tabular}
}
\caption{The accuracy of judging relative step reward.} 
\label{tab:preference}
\end{table}
\subsection{Preference Accuracy of RM}
We evaluate the quality of our RM estimated step reward by assessing its ability to determine preferences between state pairs. AgentBoard \cite{ma2024agentboard} offers a method to compute the progress rate for each state by annotating subgoals for every task. 
We create state pairs with a progress rate difference exceeding a threshold of 0.3.
Then, we calculate the accuracy of our RM in predicting preferences (Table~\ref{tab:preference}).
Despite predicting reward for each state independently, Explicit RM still demonstrates better preference judgment accuracy on most tasks compared to LLM-as-a-judge which sees pairwise states during inference.

\end{document}